\documentclass[10pt,twocolumn,letterpaper]{article}

\usepackage[pagenumbers]{wacv} % To force page numbers, e.g. for an arXiv version
\usepackage{algorithm} 
\usepackage{algorithmic}
\usepackage{graphicx}

\definecolor{wacvblue}{rgb}{0.21,0.49,0.74}
\usepackage[pagebackref,breaklinks,colorlinks,allcolors=wacvblue]{hyperref}

\title{STEC: A Reference-Free Spatio-Temporal Entropy Coverage Metric for Evaluating Sampled Video Frames}

\author{Shih-Yao Lin\\
Independent Researcher\\
{\tt\small mikeslin.research@gmail.com}
}

\begin{document}
\maketitle
\begin{abstract}
Frame sampling is a fundamental component in video understanding and video--language model pipelines,
yet evaluating the quality of sampled frames remains challenging.
Existing evaluation metrics primarily focus on perceptual quality or reconstruction fidelity,
and are not designed to assess whether a set of sampled frames adequately captures
informative and representative video content.

We propose \textbf{Spatio-Temporal Entropy Coverage (STEC)},
a simple and non-reference metric for evaluating the effectiveness of video frame sampling.
STEC builds upon \textbf{Spatio-Temporal Frame Entropy (STFE)},
which measures per-frame spatial information via entropy-based structural complexity,
and evaluates sampled frames based on their temporal coverage and redundancy.
By jointly modeling spatial information strength, temporal dispersion, and non-redundancy,
STEC provides a principled and lightweight measure of sampling quality.

Experiments on the MSR-VTT test-1k benchmark demonstrate that STEC
clearly differentiates common sampling strategies,
including random, uniform, and content-aware methods.
We further show that STEC reveals robustness patterns across individual videos
that are not captured by average performance alone,
highlighting its practical value as a general-purpose evaluation tool
for efficient video understanding.

We emphasize that STEC is not designed to predict downstream task accuracy, but to provide a task-agnostic diagnostic signal for analyzing frame sampling behavior under constrained budgets.

\end{abstract}
    
\section{Introduction}

% Video frame sampling is a critical operation in video understanding systems,
% where only a limited number of frames can be processed due to computational
% or memory constraints.
% This setting is increasingly common in modern video--language model pipelines,
% long-form video analysis, and efficient video retrieval.
% As a result, the quality of sampled frames has a significant impact
% on downstream performance.

Video frame sampling is a fundamental operation in video understanding systems, where only a limited number of frames can be processed due to computational or memory constraints. This setting is increasingly common in modern video–language model pipelines, long-form video analysis, and efficient video retrieval, making the quality of sampled frames critical to downstream performance.

Despite its importance, frame sampling is often treated as a fixed preprocessing step. Most video understanding pipelines focus on optimizing model architectures or learning objectives, leaving the quality of temporal coverage induced by sampling strategies largely unexamined and entangled with downstream tasks and models.

Moreover, there is no unified, task-agnostic metric for evaluating sparse frame sampling quality. Existing metrics mainly focus on perceptual fidelity or reconstruction and fail to assess informativeness, temporal coverage, and redundancy, making principled and task-independent comparison difficult.

% Existing metrics target perceptual quality or reconstruction fidelity and do not explicitly assess whether sampled frames capture informative content, cover the temporal extent of a video, or avoid redundancy. This limitation makes principled and task-independent comparison of sampling strategies difficult.

% Moreover, no unified, task-agnostic metric exists for evaluating sparse frame sampling quality.
% Existing metrics focus on perceptual fidelity or reconstruction and fail to assess informativeness, temporal coverage, and redundancy, making principled and task-independent comparison difficult.

In this work, we propose \textbf{Spatio-Temporal Entropy Coverage (STEC)},
a non-reference metric for evaluating video frame sampling.
STEC builds upon \textbf{Spatio-Temporal Frame Entropy (STFE)},
which quantifies per-frame spatial information using entropy-based structural complexity.
To assess a set of sampled frames, STEC further incorporates
temporal dispersion over the video timeline
and a redundancy penalty based on visual similarity.
The resulting metric jointly reflects spatial informativeness,
temporal coverage, and non-redundancy of sampled frames.

We evaluate STEC on the MSR-VTT~\cite{msrvtt} test-1k benchmark
across a range of common sampling strategies,
including random sampling, uniform sampling,
and content-aware methods.
Our results show that STEC consistently differentiates these strategies
and reveals robustness patterns at the per-video level,
providing insights that are not captured by average scores alone.
These findings suggest that STEC can serve as a practical evaluation tool
for analyzing frame sampling behavior in efficient video understanding pipelines.

Our contributions are three-fold:
\begin{itemize} 
\item  We introduce STEC, a non-reference, task-agnostic metric to evaluate arbitrary sampled frame sets, decoupled from the sampling mechanism.
\item  We propose STFE to quantify per-frame spatial information
using entropy-based structural complexity.
\item  We conduct a systematic evaluation on MSR-VTT
demonstrating that STEC effectively differentiates
common sampling strategies and reveals per-video robustness patterns.
\end{itemize}

\section{Related Work}

\subsection{STEC as an Evaluation Metric}
We emphasize that STEC does not propose a new frame sampling strategy.
Instead, it is designed as a task-agnostic evaluation metric that assesses
the quality of an arbitrary set of sampled frames, independent of how the
frames are selected.
This distinction is particularly important in modern video understanding
pipelines, where sampling decisions are often tightly coupled with model
architectures, attention mechanisms, or learned policies.
By decoupling evaluation from sampling, STEC enables fair and interpretable
comparison across heterogeneous frame selection methods.

\subsection{Frame Selection for Video Understanding and Video-LLMs}
Recent work has proposed increasingly sophisticated frame selection strategies
to reduce computational cost while preserving semantic content.
Adaptive Keyframe Sampling (AKS) dynamically selects frames for long-video
understanding under token budgets~\cite{tang2025aks}.
Hu \emph{et al.} propose an M-LLM-based frame selection module trained with
spatial and temporal supervision, which is then applied to frozen video
language models~\cite{hu2025mllm}.
KeyVideoLLM leverages CLIP-based keyframe selection to construct higher-quality
frame--text pairs for VideoLLM training~\cite{liang2024keyvideollm}.
% More recently, Logic-in-Frames introduces semantic-logical reasoning to identify
% critical frames for long video understanding~\cite{guo2025logicinframes},
% while prompting-based and zero-shot approaches use large language models
% to guide frame selection without task-specific training~\cite{wu2025contextual}.
More recently, Logic-in-Frames applies semantic–logical reasoning to identify critical frames in long videos~\cite{guo2025logicinframes}, while prompting-based and zero-shot methods leverage large language models to guide frame selection without task-specific training~\cite{wu2025contextual}.

These methods focus on improving \emph{how} frames are selected.
STEC is orthogonal: it evaluates \emph{what} is selected, and can be applied
to the outputs of any of these methods to assess informativeness, temporal
coverage, and redundancy.
We note that our experimental evaluation focuses on representative
random, uniform, and content-aware sampling strategies.
Evaluating STEC on LLM-guided or attention-based frame selection
is left for future work.
Our goal is not to exhaustively benchmark sampling methods,
but to validate STEC across representative and widely used strategies.

\subsection{Video Summarization and Subset Selection}
Video summarization shares related goals with frame sampling, aiming to select
a subset of frames or segments that best represent a video.
Dual video summarization methods jointly optimize visual and linguistic
consistency to produce semantically meaningful summaries~\cite{hu2023dual}.
% Training-free approaches such as C2FVS-DPP fuse appearance, motion, and
% semantic cues and apply determinantal point processes to encourage diversity
% and representativeness~\cite{smith2025c2fvs}.
Earlier work on submodular optimization and DPPs provides principled tools
for modeling importance and diversity~\cite{gygli2015submodular, gong2014seqdpp}.

Unlike summarization, which typically targets dense summaries and often relies on reference-based evaluation, STEC is designed for sparse frame sampling under
tight budgets and does not require human reference summaries.

\subsection{Reference-Free Evaluation of Video Summaries}
The limitations of reference-based evaluation for video summarization have
been well documented~\cite{otani2019rethinking}.
Recent work has therefore explored reference-free evaluation metrics.
QEVA evaluates narrative video summaries by generating multimodal questions
and assessing coverage, factuality, and temporal coherence without relying
on human references~\cite{jung2025qeva}.
While QEVA targets summary-level evaluation via question answering,
STEC focuses on frame-level sampling quality and provides a lightweight,
non-reference metric tailored to video understanding pipelines.

\subsection{Entropy-Based Keyframe Analysis}
Entropy has long been used as an interpretable proxy for visual information
content in keyframe extraction.
Early methods use entropy differences or relative entropy between frames
to identify representative keyframes~\cite{mentzelopoulos2004entropy,
guo2016relativeentropy}.
STEC builds upon this intuition but generalizes it to a set-level evaluation
that explicitly models temporal coverage and redundancy, which are not
addressed by entropy-only approaches.

\section{Method}

\subsection{Problem Setup}
Given a video $V$ consisting of $N$ frames $\{f_1, f_2, \dots, f_N\}$,
a sampler selects a subset of $K$ frames
$\mathcal{S} = \{f_{i_1}, f_{i_2}, \dots, f_{i_K}\}$ with $K \ll N$.
Our goal is to evaluate how informative and representative $\mathcal{S}$
is with respect to $V$ without requiring ground-truth annotations or
task-specific supervision.

\subsection{Spatio-Temporal Frame Entropy}

% \paragraph{Spatial Entropy.}
% To quantify spatial information, we compute an entropy-based measure over
% local structural responses.
% For each frame $f_i$, we first apply a Laplacian operator to obtain a
% high-frequency response map $L(f_i)$, which emphasizes local structure
% (e.g., edges and textures) while suppressing low-frequency appearance.
% The Laplacian serves as a \emph{differential operator} and does not by
% itself define an information measure.

% We therefore compute Shannon entropy over local neighborhoods of the
% Laplacian map.
% Let $\Omega$ denote the set of all pixel locations in the frame.
% For each pixel $x \in \Omega$, define the neighborhood
% $\mathcal{N}_r(x) = \{y \in \Omega \mid \|y-x\|_2 \le r\}$ with radius $r$.
% We compute entropy over the empirical distribution of values in
% $L(f_i)[\mathcal{N}_r(x)]$ and average across the spatial domain:
% \begin{equation}
% E_s(f_i) =
% \frac{1}{|\Omega|}
% \sum_{x \in \Omega}
% H\!\left(L(f_i)[\mathcal{N}_r(x)]\right).
% \label{eq:spatial_entropy}
% \end{equation}
% This formulation captures the diversity and unpredictability of local
% structural responses rather than merely their magnitude, and spatial
% averaging yields stable frame-level estimates.

\paragraph{Spatial Entropy.}
To quantify spatial information, we measure the diversity of local
structural responses using entropy.
For each frame $f_i$, we first convert the frame to grayscale and apply a
Laplacian operator to obtain a high-frequency response map $L(f_i)$,
which emphasizes local structure such as edges and textures while
suppressing low-frequency appearance.
The Laplacian acts as a differential operator and does not itself define
an information measure.

We therefore compute Shannon entropy over local neighborhoods of the
Laplacian response.
Let $\Omega$ denote the set of all pixel locations in the frame.
For each pixel $x \in \Omega$, we define a local neighborhood
$\mathcal{N}_r(x) = \{y \in \Omega \mid \|y-x\|_2 \le r\}$ with radius $r$.
The Laplacian responses within each neighborhood are quantized into a
fixed number of bins to form an empirical distribution, from which
Shannon entropy is computed.
The spatial entropy of frame $f_i$ is then obtained by averaging over all
pixel locations:
\begin{equation}
E_s(f_i) =
\frac{1}{|\Omega|}
\sum_{x \in \Omega}
H\!\left(L(f_i)[\mathcal{N}_r(x)]\right).
\label{eq:spatial_entropy}
\end{equation}
This formulation captures the unpredictability and richness of local
structural patterns rather than their raw magnitude, and spatial
averaging yields stable frame-level estimates.

% \paragraph{Temporal Entropy.}
% To capture temporal dispersion of sampled frames along the video timeline,
% we define normalized temporal positions $\tau_j = i_j/(N-1)$ for each
% $f_{i_j} \in \mathcal{S}$.
% We discretize $\{\tau_j\}$ into $B$ temporal bins and compute a normalized
% entropy:
% \begin{equation}
% E_t(\mathcal{S}) =
% - \frac{\sum_{b=1}^{B} p_b \log p_b}{\log B},
% \label{eq:temporal_entropy}
% \end{equation}
% where $p_b$ is the fraction of sampled frames falling into bin $b$.
% The normalization by $\log B$ ensures $E_t(\mathcal{S}) \in [0,1]$.

\paragraph{Temporal Entropy.}
To measure how sampled frames are distributed along the video timeline,
we consider their normalized temporal positions.
Given a video with $N$ frames indexed from $1$ to $N$, we define the
normalized temporal position of each sampled frame $f_{i_j} \in
\mathcal{S}$ as
\begin{equation}
\tau_j = \frac{i_j - 1}{N - 1},
\end{equation}
so that $\tau_j \in [0,1]$.
We discretize $\{\tau_j\}_{j=1}^{K}$ into $B$ temporal bins and compute a
normalized Shannon entropy:
\begin{equation}
E_t(\mathcal{S}) =
- \frac{\sum_{b=1}^{B} p_b \log p_b}{\log B},
\label{eq:temporal_entropy}
\end{equation}
where $p_b$ denotes the fraction of sampled frames falling into bin $b$.
Following standard convention, we define $0 \log 0 = 0$ for empty bins.
The normalization by $\log B$ ensures $E_t(\mathcal{S}) \in [0,1]$.

\subsection{Temporal Coverage and Redundancy}

\paragraph{Temporal Coverage (Span).}
Temporal entropy measures the evenness of dispersion, but does not
guarantee global coverage.
We therefore define a span factor:
\begin{equation}
C_t(\mathcal{S}) = \frac{\max_j i_j - \min_j i_j}{N-1},
\label{eq:temporal_span}
\end{equation}
and combine the two as
\begin{equation}
T(\mathcal{S}) = E_t(\mathcal{S}) \cdot C_t(\mathcal{S}).
\label{eq:temporal_coverage}
\end{equation}

\paragraph{Non-redundancy.}
We assume sampled indices are sorted in ascending temporal order,
i.e., $i_1 < i_2 < \dots < i_K$, and $K \ge 2$.
To penalize redundant selections, we compute cosine similarity between
HSV color histograms of temporally adjacent sampled frames and define
\begin{equation}
R(\mathcal{S}) =
1 - \frac{1}{K-1}\sum_{j=1}^{K-1} \mathrm{sim}(f_{i_j}, f_{i_{j+1}}),
\label{eq:redundancy}
\end{equation}
where $K = |\mathcal{S}|$ and higher $R(\mathcal{S})$ indicates lower
redundancy. We use $\ell_1$-normalized nonnegative HSV histograms, so
$\mathrm{sim}(\cdot,\cdot)\in[0,1]$ and $R(\mathcal{S})\in[0,1]$.

\subsection{Spatio-Temporal Entropy Coverage (STEC)}
We define the final Spatio-Temporal Entropy Coverage (STEC) score as the
product of spatial information strength, temporal coverage, and
non-redundancy:
\begin{equation}
\mathrm{STEC}(\mathcal{S}) =
\left(
\frac{1}{K}\sum_{j=1}^{K} E_s(f_{i_j})
\right)\cdot T(\mathcal{S}) \cdot R(\mathcal{S}),
\label{eq:stec}
\end{equation}
where $\mathcal{S}=\{f_{i_1},\dots,f_{i_K}\}$ and $K=|\mathcal{S}|$.
The multiplicative formulation enforces that high STEC scores are achieved
only when all three criteria are simultaneously satisfied, reflecting
the joint importance of spatial informativeness, temporal coverage, and
non-redundancy.

\begin{figure}[t]
    \centering
    \begin{subfigure}[t]{\linewidth}
        \centering
        \includegraphics[width=\linewidth]{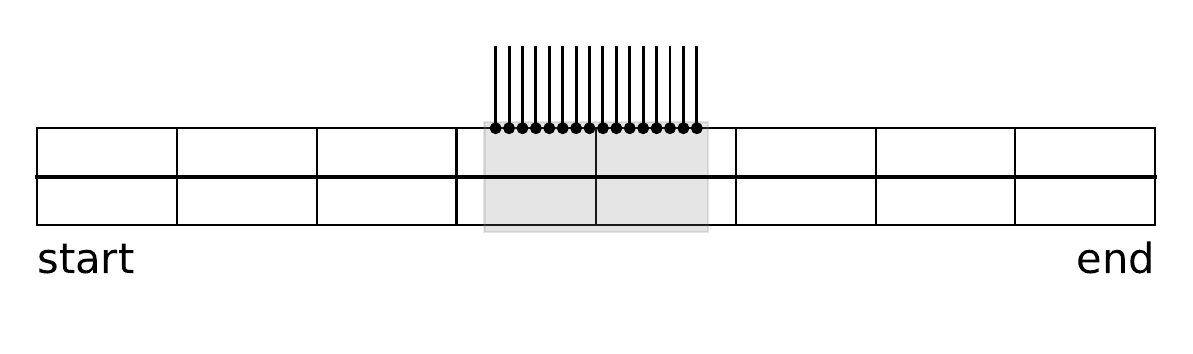}
        \vspace{-5mm}
        \caption{High temporal entropy but low temporal span
        ($E_t \approx 1,\; C_t \ll 1$).}
        
        \label{fig:entropy_low_span}
    \end{subfigure}

    % \vspace{0.6em}

    \begin{subfigure}[t]{\linewidth}
        \centering
        \includegraphics[width=\linewidth]{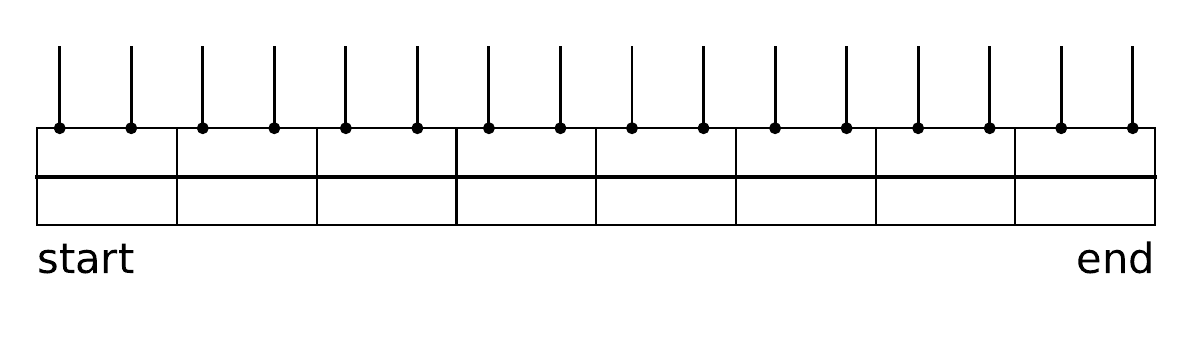}
        \vspace{-2em}
        \caption{High temporal entropy with high temporal span
        ($E_t \approx 1,\; C_t \approx 1$).}
        \label{fig:entropy_high_span}
    \end{subfigure}

    \caption{
    Temporal entropy versus temporal span.
    High temporal entropy alone does not guarantee global coverage:
    frames may be evenly distributed within a narrow temporal segment
    (top).
    Combining temporal entropy with a span-based coverage term ensures
    that sampled frames are both evenly dispersed and cover the full
    extent of the video timeline (bottom).
    }
    \label{fig:temporal_entropy_span}
    \vspace{-2mm}
\end{figure}

\paragraph{Temporal Entropy vs. Temporal Span.}
Figure~\ref{fig:temporal_entropy_span} shows a key limitation of
temporal entropy when used in isolation.
Although temporal entropy measures how evenly sampled frames are
distributed across temporal bins, it does not guarantee that these
frames cover the full extent of a video.
As shown in the top example, frames may be uniformly distributed within
a narrow temporal segment, yielding high temporal entropy but poor global
coverage.
In contrast, the bottom example demonstrates that combining temporal
entropy with a span-based coverage term ensures that sampled frames are
both evenly dispersed and span the entire video timeline.
This motivates our definition of temporal coverage as the product of
temporal entropy and temporal span.

\subsection{Discussion}
STEC is a non-reference, task-agnostic, and computationally efficient metric.
Unlike learning-based or task-specific evaluation approaches,
STEC requires no training, annotations, or pretrained models.
It also avoids reliance on optical flow, semantic representations, or supervision,
making it broadly applicable to evaluating diverse frame sampling strategies
in video understanding and video--language model pipelines.
The computational cost of STEC scales linearly with the number of sampled frames.

% \begin{figure*}[t]
% \centering
% \includegraphics[width=1\linewidth]{figures/temporal_entropy_span.pdf}
% \caption{Illustration of temporal entropy $E_t$ and temporal span $C+t$. High temporal entropy alone does not guarantee global coverage; combining with span ensures both dispersion and coverage.}
% \label{fig:temporal_entropy_span}
% \end{figure*}

\section{Experimental Setup}

\subsection{Dataset}
We evaluate the proposed STEC metric on the MSR-VTT dataset~\cite{msrvtt}, which contains diverse web videos covering a wide range of events and activities.
Following standard practice in video--language research,
we use the official test-1k split, consisting of 1,000 videos.

\subsection{Frame Sampling Methods}
We evaluate STEC on four representative frame sampling strategies:
\textbf{Random}, which selects frames uniformly at random;
\textbf{Uniform}, which samples frames at fixed temporal intervals;
\textbf{Katna}~\cite{katna}, a content-aware method based on visual importance;
and \textbf{STACFP}~\cite{keyscore}, a clustering-based strategy that selects
spatio-temporally diverse frame proposals.

For all methods, the number of sampled frames is fixed to $K=16$.
We use the STACFP frame proposal module as introduced in~\cite{keyscore}.
All samplers are evaluated under identical settings to ensure a fair comparison.

\subsection{Evaluation Protocol}
For each video, sampled frames are extracted independently using each
sampling strategy and evaluated using the proposed STEC metric.
Spatial entropy is computed from a Laplacian-based complexity map
followed by local entropy estimation.
Temporal entropy is computed by discretizing normalized frame positions
into a fixed number of temporal bins.
Redundancy is measured using cosine similarity between color histograms
of temporally adjacent sampled frames.
All hyperparameters are kept fixed across experiments,
including the entropy neighborhood size, the number of temporal bins,
and histogram resolutions, to ensure consistent evaluation across samplers.

\subsection{Implementation Details}
For spatial entropy computation, each frame is first converted to grayscale
and processed using a $3\times3$ Laplacian operator.
Concretely, we normalize the Laplacian response map to 8-bit and compute
local Shannon entropy using a disk-shaped structuring element
(radius $r$) via a rank-based entropy filter, which implements Eq.~(1)
with implicit binning over 256 gray levels.

All experiments are conducted without any learning or fine-tuning.
STEC is computed directly from sampled frames and their temporal indices.
Unless otherwise specified, we use $K=16$ sampled frames,
a fixed number of temporal bins $B=8$, and a local entropy neighborhood
radius of $6$ pixels.
All hyperparameters are kept fixed across experiments to ensure fair comparison.

\begin{table}[t]
\small
\centering
\caption{
Comparison of frame sampling strategies on MSR-VTT test-1k
using the proposed Spatio-Temporal Entropy Coverage (STEC) metric.
Higher values indicate better sampling quality / coverage quality.
}
\setlength{\tabcolsep}{6pt}
\begin{tabular}{lcccc}
\toprule
\textbf{Sampler} 
& \textbf{S} $\uparrow$ 
& \textbf{T} $\uparrow$ 
& \textbf{R} $\uparrow$ 
& \textbf{STEC} $\uparrow$ \\
\midrule
Random   
& 1.87 
& 0.82 
& 0.21 
& 0.306 \\

Uniform  
& 2.41 
& \textbf{1.00} 
& 0.19 
& 0.463 \\

Katna~\cite{katna}
& \underline{2.95} 
& 0.91 
& \underline{0.25} 
& \textbf{0.663} \\

STACFP~\cite{keyscore}
& \textbf{3.02} 
& 0.88 
& \textbf{0.27} 
& \underline{0.658} \\
\bottomrule
\end{tabular}
\label{tab:stec_main}
\end{table}

\begin{table}[t]
\small
\centering
\caption{
Number of videos (out of 1,000) for which each sampling strategy
achieves the highest STEC score on MSR-VTT test-1k.
}
\begin{tabular}{lc}
\toprule
\textbf{Sampler} & \textbf{\#Videos with Highest STEC} \\
\midrule
Random   & 7 \\
Uniform  & 91 \\
Katna~\cite{katna}    & 421 \\
STACFP~\cite{keyscore}& \textbf{481} \\
\bottomrule
\end{tabular}
\label{tab:stec_wins}
\end{table}

\section{Results and Analysis}

\subsection{Quantitative Results}
As a sanity check, STEC assigns the lowest scores to random sampling
and higher scores to more structured sampling strategies.
Table~\ref{tab:stec_main} shows the average spatial information score ($S$),
temporal coverage score ($T$), redundancy score ($R$),
and the STEC score on MSR-VTT test-1k split.

Uniform sampling achieves perfect temporal coverage by construction,
but suffers from lower spatial information and higher redundancy,
resulting in a moderate STEC score.
Random sampling performs poorly across all metrics,
highlighting the difficulty of capturing informative content without structure.

Content-aware sampling methods, including Katna~\cite{katna} and STACFP~\cite{keyscore},
substantially improve spatial information and non-redundancy,
leading to significantly higher STEC scores.
Katna achieves the highest average STEC score,
while STACFP attains a comparable performance.
This suggests that while Katna excels on average,
STACFP exhibits more consistent performance across individual videos.
This contrast motivates a per-video robustness analysis,
as average performance alone does not capture consistency
across diverse videos.

\begin{figure}[t]
    \centering
    \includegraphics[width=\linewidth]{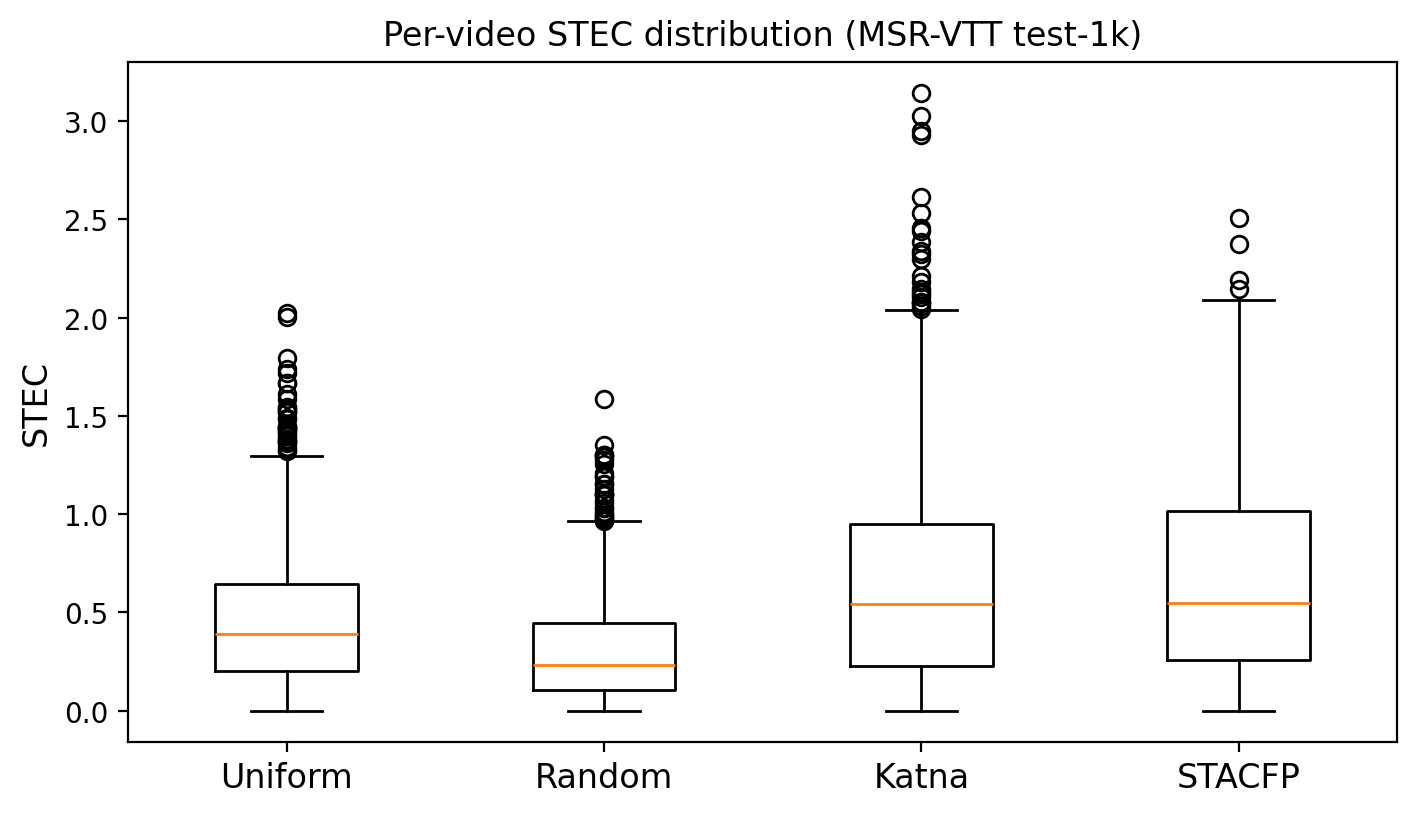}
    \caption{\textbf{Distribution of per-video STEC scores on MSR-VTT test-1k.}
    Boxplots summarize STEC values across individual videos for different
    sampling strategies.
    The median and interquartile range represent typical performance,
    while outliers correspond to rare extreme cases.
    Content-aware methods (Katna and STACFP) shift the overall distribution
    toward higher STEC values, indicating improved robustness across diverse videos, rather than performance dominated by a small number of outlier cases.
    Notably, STACFP exhibits a more compact distribution with fewer extreme failures, indicating stronger per-video robustness compared to methods with higher variance.
    }
    \label{fig:stec_distribution}
    \vspace{-2mm}
\end{figure}

\begin{figure*}[t]
    \centering
    \includegraphics[width=1.0\linewidth]{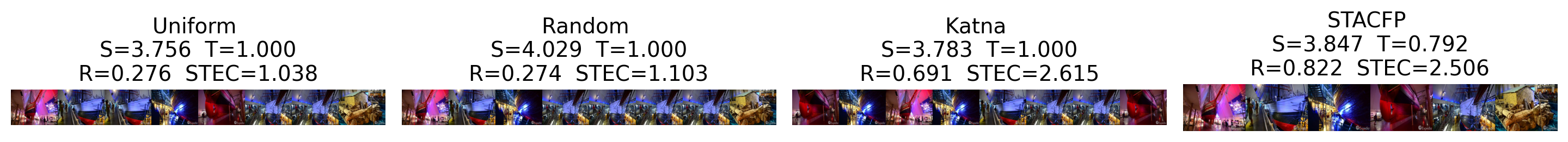}\\[4pt]
    \includegraphics[width=1.0\linewidth]{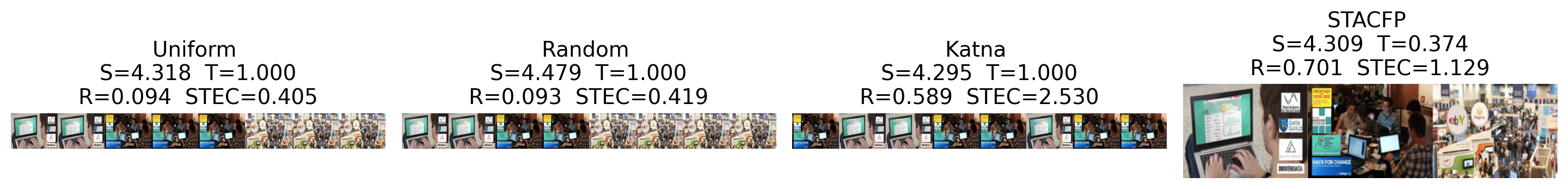}
\caption{\textbf{Qualitative analysis of frame sampling behavior using STEC.}
Two representative MSR-VTT videos illustrate the coverage--redundancy trade-off captured by STEC.
\textbf{Top:} For a video with diverse scenes, Uniform and Random achieve full temporal coverage but select many redundant frames, resulting in low STEC.
Content-aware methods improve diversity: STACFP focuses on informative segments with higher non-redundancy, while Katna maintains full coverage and achieves the highest STEC.
\textbf{Bottom:} For a video dominated by repetitive content, Uniform and Random repeatedly select near-duplicate frames, leading to very low non-redundancy.
Content-aware methods explicitly reduce redundancy; STACFP emphasizes diversity with reduced coverage, whereas Katna balances coverage and diversity and yields the highest STEC.
%These examples illustrate how STEC disentangles temporal coverage from redundancy, providing diagnostic signals beyond coverage-only measures.
}

    \label{fig:qualitative_sampling}
\end{figure*}

\subsection{Per-Video Robustness Beyond Average Performance}
Beyond average performance, we analyze robustness at the per-video level.
Table~\ref{tab:stec_wins} reports how often each sampling strategy achieves
the highest STEC score across individual videos.
STACFP attains the highest STEC on the largest portion of videos,
indicating more consistent performance across diverse content.

To further examine robustness, Figure~\ref{fig:stec_distribution}
shows the distribution of per-video STEC scores.
Uniform and Random sampling concentrate at low STEC values,
reflecting systematic redundancy despite full temporal coverage.
In contrast, content-aware methods shift the median and interquartile range
toward higher values.
Katna occasionally achieves very high STEC on specific videos,
whereas STACFP exhibits a more compact distribution with consistently
strong performance.

Together, these results show that STEC reveals per-video
robustness patterns that are not evident from average
statistics alone.

\paragraph{From STEC Evaluation to Downstream Retrieval.}
STEC is designed as a simple, task-agnostic metric for assessing the
quality of sampled frame sets, without optimizing for any specific
downstream task.
A practical question is whether frame sampling strategies that score well
under STEC still work reasonably well when used in real
video--language retrieval systems.

To study this, we run a retrieval experiment on MSR-VTT using a fixed
video--text encoder and the same retrieval setup across all methods.
We compare random sampling, uniform sampling, appearance-based clustering
(Katna)~\cite{katna}, and STACFP~\cite{keyscore}.
The model, encoder, and retrieval protocol are kept identical, and only
the selected frame sets differ, with a fixed sampling budget of $K=16$.

Table~\ref{tab:retrieval_msr_vtt} shows that sampling strategies with
better spatio-temporal coverage under STEC generally do not degrade
retrieval performance under a fixed frame budget.
In particular, STACFP achieves the strongest retrieval results while also
obtaining high STEC scores, suggesting that frame sets with good temporal coverage and low redundancy remain compatible with a frozen retrieval model.
At the same time, higher STEC scores do not always translate to higher
retrieval accuracy, indicating that STEC is not a predictor of downstream
performance.

Overall, this experiment does not claim that STEC can replace task-level
evaluation.
Instead, it demonstrates that STEC provides a reasonable,
task-independent way to screen frame sampling strategies and identify
selections that are unlikely to degrade downstream retrieval when compute
or sampling budgets are limited.

% new added
\begin{table}[t]
\centering
\caption{Downstream retrieval performance on MSR-VTT using different frame sampling strategies under a fixed video--text encoder.
This experiment examines whether higher STEC scores correspond to preserved downstream retrieval performance.
T2V/V2T: Recall@1 (\%).}
\label{tab:retrieval_msr_vtt}
\begin{tabular}{lcc}
\toprule
\textbf{Frame Sampler} & \textbf{T2V} & \textbf{V2T} \\
\midrule
Random               & 49.2 & 46.1 \\
Uniform                & 50.6 & 48.0 \\
Katna~\cite{katna}  & 49.9 & 46.8 \\
% SACFP (LMSKE~\cite{LMSKE39}) & 50.2 & 47.4 \\
STACFP~\cite{keyscore}                      & \textbf{50.9} & \textbf{48.8} \\
\bottomrule
% \vspace{-2em}
\end{tabular}
\end{table}

\subsection{Qualitative Analysis of Sampling Behavior}
While aggregate statistics compare sampling strategies on average, they do not reveal the underlying reasons for their behavior on individual videos.
We therefore provide a qualitative analysis to visualize the sampling patterns and failure modes captured by STEC.

Figure~\ref{fig:qualitative_sampling} presents two representative MSR-VTT videos illustrating distinct sampling scenarios.
In the first case (top), Uniform and Random sampling achieve perfect temporal coverage but select many visually similar frames, resulting in low non-redundancy and low STEC.
STACFP substantially increases diversity by focusing on informative segments, leading to a much higher non-redundancy score, albeit with reduced coverage.
Katna maintains full coverage while reducing redundancy, achieving the highest STEC on this video.
This example highlights that temporal coverage alone is insufficient to characterize sampling quality.

The second case (bottom) shows a video with highly repetitive visual content.
Although Uniform and Random again achieve perfect coverage, they repeatedly sample near-duplicate frames, resulting in extremely low non-redundancy.
Content-aware methods explicitly mitigate redundancy by selecting more diverse frames.
STACFP concentrates on a subset of informative segments, while Katna balances coverage and diversity, yielding the highest STEC.
Together, these examples demonstrate that STEC provides a diagnostic view of coverage--redundancy trade-offs that are invisible to temporal coverage alone.

\section{Discussion}
\label{sec:discussion}

Taken together, these results demonstrate that STEC effectively captures key aspects of frame sampling quality, including spatial informativeness, temporal coverage, and redundancy.
The proposed metric clearly differentiates between naive, uniform, and content-aware sampling strategies, highlighting its usefulness as a general-purpose, non-reference evaluation tool for analyzing sampled frame sets across diverse video pipelines. 
STEC is intended for comparative evaluation of frame sampling strategies under fixed budgets, rather than absolute assessment of video understanding performance.

\paragraph{Relation to Downstream Tasks.}
Although STEC is designed as a task-agnostic metric rather than a predictor of absolute downstream task accuracy, an important practical question is whether higher STEC scores correspond to more effective downstream usage.
As shown in Table~\ref{tab:retrieval_msr_vtt}, frame sampling strategies
with strong spatio-temporal coverage under STEC generally remain
competitive in downstream retrieval under a fixed video--text encoder
and a fixed frame budget.

This observation suggests that STEC captures spatio-temporal properties of frame sets that are broadly aligned with the requirements of practical video--language applications.
We emphasize that this analysis demonstrates consistency rather than causality, and STEC should be viewed as a complementary evaluation signal rather than a replacement for task-specific performance metrics.

\paragraph{Limitations of Spatial Entropy.}
The spatial entropy component of STEC measures information richness based on low-level visual statistics and does not explicitly account for high-level semantic importance.
As a result, scenes that are semantically critical but visually homogeneous may receive lower spatial entropy scores.
This design choice allows STEC to remain lightweight and training-free, but also highlights an important limitation.
Future work could incorporate semantic-aware alternatives, such as entropy computed over pretrained visual--language embeddings or object-conditioned representations, to better capture task-relevant semantics.

\paragraph{Sensitivity to Hyperparameters.}
In this work, we focus on demonstrating the effectiveness of STEC under a fixed and representative set of hyperparameters.
While our experiments indicate that STEC provides meaningful and interpretable comparisons across different sampling strategies, a systematic sensitivity analysis with respect to hyperparameters such as the spatial entropy neighborhood size, temporal binning scheme, and sampling budget remains an important direction for future work.
We plan to examine the stability of STEC-induced rankings under varying configurations and across datasets, with the goal of further characterizing the robustness of the proposed metric.

\paragraph{Design Choices and Robustness.}
STEC employs a multiplicative formulation to jointly model spatial information, temporal coverage, and redundancy.
This design enforces balanced spatio-temporal coverage by penalizing degenerate solutions that optimize only a single component.
% Our sensitivity analysis indicates that the relative ranking induced by STEC remains stable across a wide range of hyperparameter settings, including spatial window sizes, temporal binning schemes, and sampling budgets.
% These results suggest that STEC is robust to reasonable parameter choices and suitable for comparative evaluation of diverse frame sampling strategies.
The results suggest that STEC provides consistent comparative signals under the evaluated parameter settings, enabling interpretable comparison among diverse frame sampling approaches.

\paragraph{Scope and Applicability.}
We primarily evaluate STEC on short- to medium-length videos commonly used in video--language benchmarks.
While the formulation naturally extends to longer videos, its behavior under extreme low-motion or highly repetitive content warrants further investigation.
We view STEC as a first step toward principled, non-reference evaluation of frame sampling quality, and hope it will encourage future research on task-consistent and semantically informed evaluation metrics.

\section{Conclusion}

We introduced Spatio-Temporal Entropy Coverage (STEC),
a simple and non-reference metric for evaluating the effectiveness
of video frame sampling.
By jointly modeling spatial information, temporal coverage,
and redundancy, STEC provides a principled and lightweight evaluation
of how well sampled frames represent video content.
Experiments on MSR-VTT demonstrate that STEC clearly differentiates
common sampling strategies and reveals robustness patterns
that are not captured by average performance alone.
We believe STEC provides a practical evaluation tool
for analyzing frame sampling behavior
in efficient video understanding pipelines.
%
% \paragraph{Code Availability.}
% The implementation of STEC and the scripts for reproducing the experiments are publicly available at \url{https://github.com/mikeslin-embodai/STEC}.

\section*{Code Availability}

The reference implementation of STEC and all evaluation scripts
are publicly available at:
\url{https://github.com/mikeslin-ir/STEC}.

% \clearpage
{
    \small
    \bibliographystyle{ieee_fullname}
    \bibliography{stec_refs}
}

\end{document}